\documentclass[10pt,twocolumn,letterpaper]{article}

\usepackage[pagenumbers]{cvpr} % To force page numbers, e.g. for an arXiv version

\usepackage[T1]{fontenc}
\usepackage{multirow}
\usepackage{graphicx}
\usepackage{amsmath}
\usepackage{amssymb}
\usepackage{booktabs}
\usepackage{multirow}
\usepackage{algorithm}
\usepackage{algpseudocode}

\definecolor{cvprblue}{rgb}{0.21,0.49,0.74}
\usepackage[pagebackref,breaklinks,colorlinks,allcolors=cvprblue]{hyperref}

\def\paperID{7915} % *** Enter the Paper ID here
\def\confName{ICCV}
\def\confYear{2025}

\title{Make the Most of Everything: Further Considerations on Disrupting Diffusion-based Customization}

\author{Long Tang, Dengpan Ye*, Sirun Chen, Xiuwen Shi, Yunna Lv, Ziyi Liu\\
School of  Cyber Science and Engineering, Wuhan University, Wuhan 430072, China\\
{\tt\small \{l\_tang,yedp,2023202210171,2024282210162,lvyunna,ziyi\_liu\}@whu.edu.cn}
}
\begin{document}
\maketitle
\begin{abstract}
The fine-tuning technique for text-to-image diffusion models facilitates image customization but risks privacy breaches and opinion manipulation. Current research focuses on prompt- or image-level adversarial attacks for anti-customization, yet it overlooks the correlation between these two levels and the relationship between internal modules and inputs. This hinders anti-customization performance in practical threat scenarios. We propose \textbf{D}ual \textbf{A}nti-\textbf{Diff}usion (DADiff), a two-stage adversarial attack targeting diffusion customization, which, for the first time, integrates the adversarial prompt-level attack into the generation process of image-level adversarial examples. In stage 1, we generate prompt-level adversarial vectors to guide the subsequent image-level attack. In stage 2, besides conducting the end-to-end attack on the UNet model, we disrupt its self- and cross-attention modules, aiming to break the correlations between image pixels and align the cross-attention results computed using instance prompts and adversarial prompt vectors within the images. Furthermore, we introduce a local random timestep gradient ensemble strategy, which updates adversarial perturbations by integrating random gradients from multiple segmented timesets. Experimental results on various mainstream facial datasets demonstrate 10\%-30\% improvements in cross-prompt, keyword mismatch, cross-model, and cross-mechanism anti-customization with DADiff compared to existing methods.
\end{abstract}    
\section{Introduction}
\label{sec:intro}

\begin{figure}[tb]
    \centering
    \includegraphics[width=1.0\linewidth]{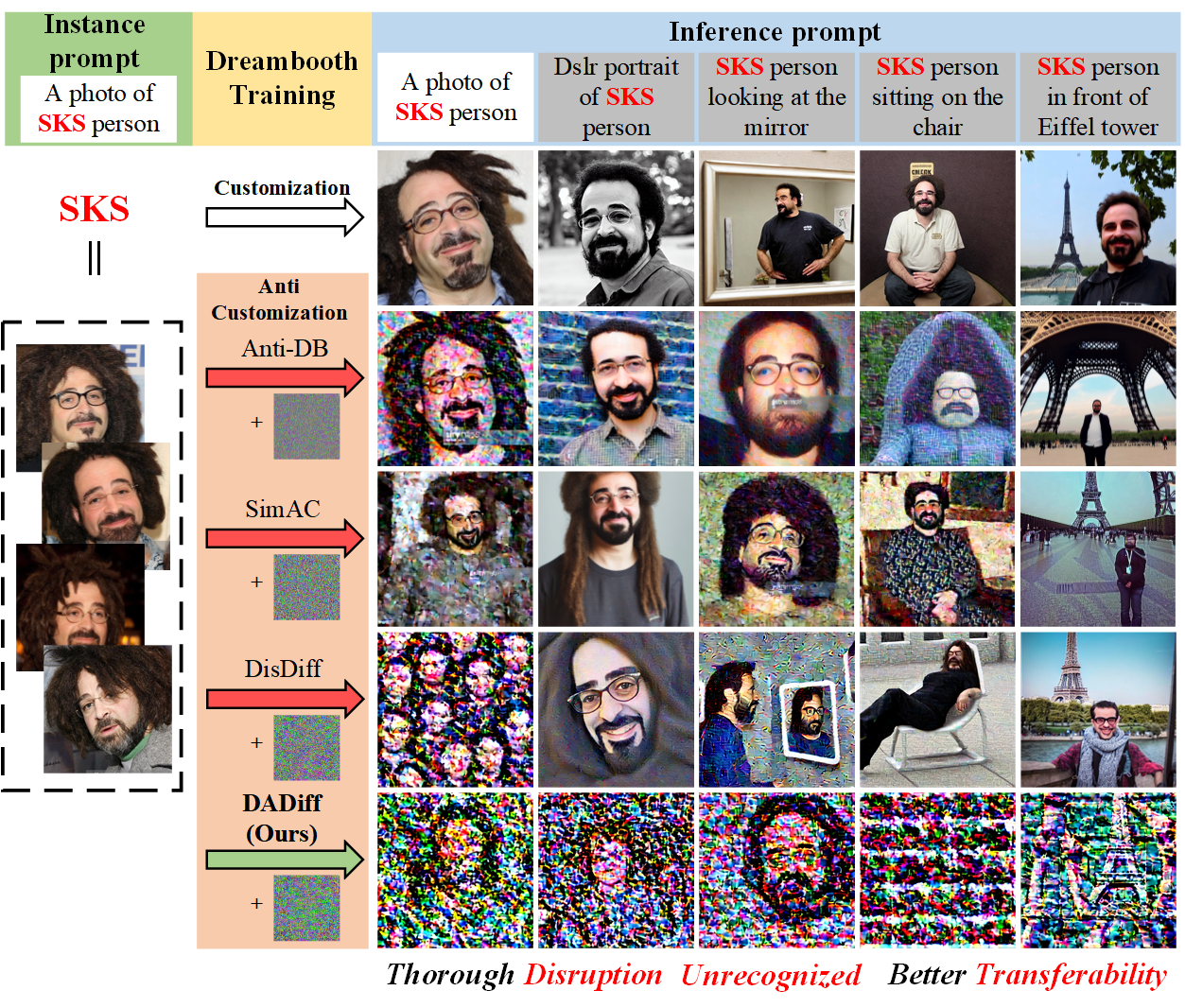}
    \caption{Comparison of vanilla Dreambooth customization, SOTA anti-customization methods, and proposed DADiff. Inference prompts in the grey background are unknown prompts for anti-customization. DADiff achieves a more thorough disruption, making generated images hard to recognize, and has better transferability across different black-box inference prompts.}
\label{fig1}
\end{figure}

The rapid development of diffusion models has improved image generation quality significantly, leading to fundamental innovations in image restoration, multi-modal synthesis, and image customization. Studies \cite{huang2024diffusion} have shown that large pre-trained models have low intrinsic dimensions, meaning that fine-tuning in low-dimensional parameter space can yield results similar to fine-tuning the entire model. Therefore, model customization through fine-tuning techniques is highly favored, allowing for the execution of specific high-quality generation tasks by training with minimal image inputs. Typical fine-tuning techniques for text-to-image generation like Dreambooth \cite{ruiz2023dreambooth}, LoRA \cite{hu2022lora}, and Textual Inversion \cite{gal2022textual}, require only a few images of a specific individual to obtain a customized diffusion model capable of generating images of that individual, as shown in the first row of Fig. \ref{fig1}. These technologies have significantly reduced the cost of artistic creation and advanced the development of artificial intelligence generation techniques.

However, malicious users can generate massive fake images by obtaining personal images of victims and using customized fine-tuning techniques, leading to privacy breaches. These forged images, exploited for profit in sectors like the adult industry, fraud, and political propaganda, cause immense harm to the victims. Unlike traditional Deepfakes relying on Generative Adversarial Networks (GANs) for image editing or face swapping, Deepfakes based on fine-tuning diffusion models utilize victim images for model training. This allows customizing keywords to refer to the victim and the generation of diverse fake images using different prompts. Furthermore, the image generation mechanism of diffusion models differs from that of GANs-based Deepfakes, making traditional active defense methods challenging to apply directly. To address this issue and prevent images from being misused for fine-tuning diffusion models, research on protective measures is urgently needed.

% A prevalent protective method is utilizing adversarial examples to introduce imperceptible perturbations into protected images. Malicious users can just obtain low-quality fake images using the diffusion model fine-tuned with protected images. While there have been some related studies, these methods either target the non-core module of Stable Diffusion (SD) models \cite{salman2023raising,ye2024duaw} or employ only a single random timestep iteratively to attack the UNet in SD models. Some studies have considered prompt augmentation at the embedding level, lacking exploration into deeper and more complex prompt embedding patterns. These limitations have led to sub-optimal performance in addressing scenarios involving cross-keyword or cross-prompt situations, making it challenging to effectively mitigate the risks of model customization in real-world settings. The inputs of the diffusion model encompass three crucial elements: image, prompt, and timesteps. We believe that fully leveraging the input information can significantly enhance the effectiveness of adversarial-based anti-customization.

A prevalent protective method is utilizing adversarial examples to introduce imperceptible perturbations into protected images. Malicious users can just obtain low-quality fake images using the diffusion model fine-tuned with protected images. Related studies either target the non-core module of Stable Diffusion (SD) models \cite{salman2023raising,ye2024duaw} or employ only a single random timestep iteratively to attack the UNet in SD models \cite{liang2023adversarial,van2023anti,wang2024simac,liu2024disrupting}. Moreover, existing methods generally focus on enhancing image-level adversariality, ignoring the potential impact of prompt-level adversariality on image examples. These limitations have led to less optimal performance in addressing scenarios involving cross-keyword or cross-prompt situations, making it challenging to effectively mitigate the risks of model customization in real-world settings. The inputs of the diffusion model encompass three crucial elements: image, prompt, and timesteps. We believe that fully leveraging the input information can significantly enhance the effectiveness of adversarial-based anti-customization.

This paper proposes a two-stage adversarial attack tailored for the anti-customization of diffusion models, named \textbf{D}ual \textbf{A}nti-\textbf{Diff}usion (DADiff). DADiff explores the full input contents and internal functional modules of diffusion models, aiming to enhance the effectiveness and generalizability of image perturbations against diffusion customization. Specifically, we first conduct adversarial attacks on the prompts to generate adversarial prompt vectors, providing adversarial prior knowledge for the subsequent image-level attack. Then, during the image-level attack, beyond considering the loss of UNet output, we design loss functions for the model's internal self-attention and cross-attention modules, aiming to disrupt the attention relationships among pixels in adversarial examples and ensure that the cross-attention guided by normal prompts and adversarial prompt vectors remains aligned. Furthermore, existing methods neglect gradient correlations across timesteps. We propose the local random timestep gradient ensemble strategy, which updates adversarial examples by randomly selecting timesteps within equal interval groups and integrating adversarial gradients, thereby more effectively capturing gradient correlations in the time dimension and further enhancing the anti-customization of adversarial examples.

Our contributions are summarized as:
\begin{itemize}
\item We introduce DADiff, a two-stage adversarial attack against diffusion models, combining prompt-level and image-level attacks for the first time to enhance the effectiveness of diffusion anti-customization. The adversarial guidance in prompt vectors provides prior information for adversarial attacks in the image modal.
\item We utilize the diffusion UNet's diverse inputs and attention modules. We design attack losses on self-attention, cross-attention, and output levels to interfere with the model's pixel correlation, image-textual semantic correlation, and output prediction. Furthermore, we propose a local random timestep gradient ensemble strategy, which integrates gradients of noisy images across multiple timesteps to improve gradient optimization.
\item By conducting experiments on stable diffusion models v1.4, v1.5, and v2.1 using multiple images with different IDs from the CelebA-HQ and VGGFace2 datasets, our proposed DADiff achieves a significant improvement of approximately 10\%-30\% in multiple evaluation metrics compared to existing works, particularly in cross-prompt, cross-model, and cross-mechanism settings.
\end{itemize}
\section{Related Work}
\label{sec:related}

\subsection{Diffusion and Customization}
Diffusion models \cite{ho2020denoising} utilize a continuous Markov chain to progressively add Gaussian noise into training data and subsequently reverse this process to reconstruct data samples. Stable Diffusion \cite{rombach2022high} applies this mechanism to text-to-image synthesis, operating in latent space and incorporating cross-attention modules to accommodate diverse user-defined prompts. It has emerged as a pivotal technology that enables users to create customized visual content.
                                                                
Diffusion-based customization techniques have been proposed to tailor the generation process for specific needs. LoRA \cite{hu2022lora} fine-tunes the diffusion model with additional parameters to link image representations with descriptive prompts. Textual Inversion \cite{gal2022textual} leverages the flexible word embedding space of the text encoder to map novel concepts to pseudo-words. A notable customization method is Dreambooth \cite{ruiz2023dreambooth}, which fine-tunes a pre-trained Stable Diffusion model using just 3 to 5 images and a corresponding identifier, enabling the model to "memorize" and reproduce specific concepts in new contexts. The integration of Dreambooth with LoRA has further optimized its efficiency and inspired numerous subsequent customization projects. However, ethical considerations and safeguards are paramount to guard against the misuse of such technologies, including the generation of illicit or harmful content.

\subsection{Adversarial Attacks against Deepfakes}
As an emerging form of Deepfake technology, diffusion still faces an exploratory phase in terms of defensive measures, despite that countermeasures against traditional Deepfake models have already matured. Traditional Deepfake models are mostly constructed on the GAN framework. In this field, Yang et al. \cite{yang2021defending} exploits differentiable image transformations for robust image cloaking; Ruiz et al. \cite{ruiz2020disrupting} leverage the Projected Gradient Descent (PGD) attack to disrupt the output of GAN, thereby preventing StarGAN from producing fake images with attribute manipulations; Huang et al. \cite{huang2021initiative} propose a generative adversarial attack framework by training surrogate models to attack in a gray-box setting; UnGANable \cite{li2023unganable} prevents image editing by disrupting the image inversion process of StyleGAN; CMUA-Watermark \cite{huang2022cmua} and FOUND \cite{tang2024feature} design universal adversarial watermarks that can be effectively applied across multiple face images and successfully resist attribute editing from various GAN models. It is noteworthy that diffusion models exhibit significant differences from GANs in terms of architectural design and fake image generation mechanisms. This characteristic makes it difficult to directly transfer countermeasures designed for GANs to diffusion models.

\subsection{Adversarial Attacks against Diffusions}
The recent adversarial attacks on text-to-image diffusion models can be divided into prompt-level and image-level attacks. At the text level, Zhuang et al. \cite{zhuang2023pilot} propose a text-level adversarial character attack for the Stable Diffusion model. By minimizing the cosine similarity between the adversarial text and the normal text encoding, attackers can use just 5 additional adversarial characters to significantly change the generated images. Yang et al. \cite{yang2024mma} enhance the adversarial attack of text modality through image gradient, enabling adversarial text to circumvent the sensitive word and safety checkers of Stable Diffusion and generate prohibited images. At the image level, PhotoGuard \cite{salman2023raising} attacks both VAE encoder and UNet model and targets a gray image. AdvDM \cite{liang2023adversarial} directly uses PGD \cite{madry2018towards} to attack the UNet model to prevent Textural Inversion. Anti-Dreambooth (Anti-DB) \cite{van2023anti} uses the Alternating Surrogate and Perturbation Learning (ASPL) to approximate the real trained models and alternately performs Dreambooth training and attack. SimAC \cite{wang2024simac} further leverages a greedy algorithm to select timesteps with the highest gradient scores to update the adversarial example. DisDiff \cite{liu2024disrupting} additionally set cross-attention erasure loss to erase the keyword's attention in attacking the Dreambooth process.

However, despite attempts to leverage image gradients in prompt-level adversarial attacks, adversarial attacks at the image level have yet to establish a connection with prompt-level attacks. Additionally, while existing research has noticed the impact of timesteps, they only use a single timestep to generate gradients in one iteration, neglecting the diversity of gradients in images under different noise conditions. Therefore, there is still significant room for development in adversarial attacks on diffusion models at the image level.
\section{Preliminaries}
\label{sec:Preliminary}
\subsection{Diffusion Models}

Diffusion Models are generative models divided into a forward process (adding noise to structured data) and an inverse process (recovering or generating data by denoising).
Specifically, for a given image $x_{0}\sim q(x)$, the forward process gradually adds noise according to a pre-set noise scheduler $\{\beta_{t}:\beta_{t}\in (0, 1)\}^{T}_{t=1}$, generating a series of intermediate noisy images $\{x_1, x_2, ..., x_T\}$, until the image is completely transformed into the Gaussian noise $\epsilon\sim\textit{N}(0, \textbf{I})$. The process is formulated in Eq. (\ref{eq1}), where $\alpha_{t}=1-\beta_{t}$ and $\bar\alpha_{t}=\prod_{s=1}^{t}\alpha_s$. In previous studies of diffusion models, this process typically involves $T=1000$ steps.
\begin{equation}\label{eq1}
    x_t=\sqrt{\bar\alpha_{t}}x_0 + \sqrt{1-\bar\alpha_{t}}\epsilon.
\end{equation}

In the inverse process, the noise is gradually reduced from $x_T$, until the original image $x_0$ is finally restored. During this process, the diffusion UNet model $\epsilon_{\theta}$ typically learns how to predict the $\epsilon$ added to the noisy image $x_t$ in the forward process based on the noisy image $x_{t+1}$. For diffusion models that accept prompt inputs, to ensure that the prompt can effectively control the diversity of diffusion outputs, the model $\epsilon_{\theta}$ employs a cross-attention module to fuse image and prompt information, ultimately returning predicted sampling noise. Therefore, the loss function of training the model aims to minimize the $L_2$ distance between the model's output and the sampling Gaussian noise $\epsilon$, and the model's input includes the noisy image $x_{t+1}$ at time $t+1$, condition prompt $P$, and the current timestep $t$:
\begin{equation}\label{eq2}
    L_{cond}(\theta, x_{0},P)=\mathbb{E}_{x_{0},t,\epsilon}||\epsilon-\epsilon_{\theta}(x_{t+1},t,P)||_2^2.
\end{equation}

Note that the timestep $t$ is usually omitted in the loss function since it is a fixed and discrete random input, which determines how much noise will be added to the image in the diffusion process. 
% The main factors that truly influence the output are the noisy image and the condition.

\paragraph{Dreambooth}
We primarily introduce the fine-tuning of diffusion models using Dreambooth as an example. Dreambooth is a technique that enables the generation of new images with specific thematic features by personalized fine-tuning a pre-trained diffusion model using a small number of reference images of custom themes or objects. Assuming we have a set of photos of person \textit{[A]}, to enable the model to generate customized images related to him, we can leverage a combination of a new prompt $P_{new}=$"\textit{a photo of [A] person}" and the base prompt $P_{base}=$"\textit{a photo of person}" to design the loss function for fine-tuning the diffusion model:
\begin{equation}
\begin{split} 
    \textit{L}_{db}(\theta, x_{0}) & = \mathbb{E}_{x_{0},t,\epsilon}||\epsilon-\epsilon_{\theta}(x_{t+1},t,P_{base})||_2^2 \\
    & + \lambda||\epsilon'-\epsilon_{\theta}(x'_{t'+1},t',P_{new})||_2^2,
\end{split}
\end{equation}
where $\epsilon$ and $\epsilon'$ are Gaussian noises, $t$ and $t'$ are random timesteps, $x_{t+1}$ is the noisy image of person \textit{[A]} and $x_{t'+1}'$ is the generated image of "\textit{person}" by pre-trained $\epsilon_{\theta}$ and noised in $t'+1$ timestep. \textit{[A]} can be any custom keywords, such as "\textit{sks, asdf...}". For more details, refer to \cite{ruiz2023dreambooth}.

\subsection{Adversarial Attack-Based Anti-Customization}
We introduce generating adversarial examples for anti-customization. Given a set of images $X$ of a person \textit{[A]}, we add imperceptible perturbations $\delta$ to $X$, making it impossible for an adversary to utilize $X$ for Dreambooth training. The current optimization objective for anti-customization is to maximize the $L_2$ distance between the output of the diffusion UNet and Gaussian noise $\epsilon$:
\begin{equation}\label{eq4}
    \delta\leftarrow \mathop{\arg\max}\limits_{\delta}L_{cond}(\theta,x_{0}+\delta,P_{new}),
\end{equation}
where $x_{0}\in X$, $L_{cond}(\cdot)$ is defined in Eq. (\ref{eq2}) and $P_{new}$ is the new instance prompt. After determining the loss function, the PGD attack \cite{madry2018towards} is commonly used to update the adversarial examples, which iteratively update the perturbation through multiple signed gradients:
\begin{equation}
        \label{eq5}
        \delta_{adv}^{r+1}=clip_{\delta,\omega}(\delta_{adv}^{r}+\eta sign(\nabla_{\delta}L(\theta,x_{0}+\delta_{adv}^{r},P_{new}))),
\end{equation}
where $r\in R$ is the number of steps for the PGD attack, $\eta$ is the learning rate, and $\omega$ is the perturbation constraint, which is typically measured using the $l_\infty$ norm.

\section{Method}
\label{sec:method}
\vspace{-0.1cm}
\subsection{Overview}
\begin{figure*}[t]
    \centering
    \includegraphics[width=0.95\linewidth]{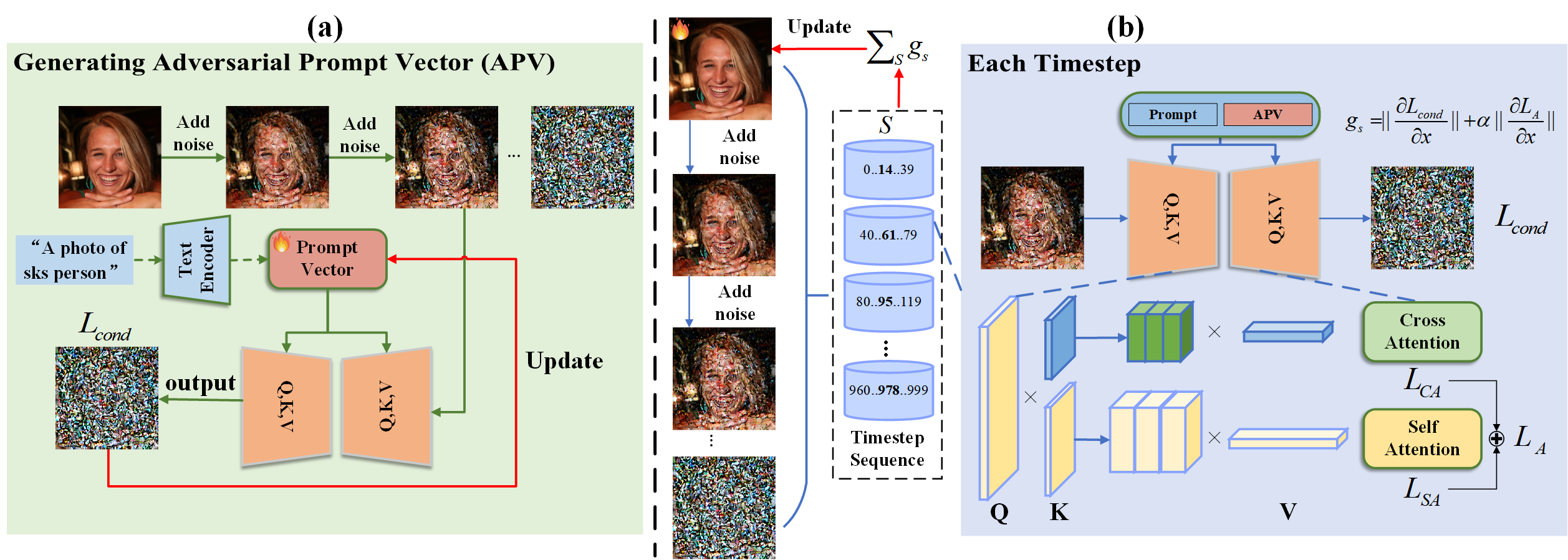}
    \caption{The pipeline of DADiff, where only the \textbf{flame} icon is updated at each step. We firstly execute stage (a) to obtain the Adversarial Prompt Vector, and then use APV and instance prompt to generate image-level adversarial examples in stage (b).}
\label{fig2}
\end{figure*}
\vspace{-0.1cm}

We display the implementation pipeline of DADiff. As shown in Fig. \ref{fig2}(a), we first use the adversarial loss calculated by Eq. (\ref{eq2}) to update the prompt-level Adversarial Prompt Vector (APV) iteratively. Subsequently, we process the image-level adversarial attack in Fig. \ref{fig2}(b). We simultaneously monitor the output of the attention module and the terminal output of the model. For gradient calculation at a certain timestep, we adopt a multidimensional strategy: on the one hand, we require the output of the UNet model to leave from Gaussian noise; On the other hand, for the self-attention module, we maximize the difference between the self-attention output of adversarial examples and that of clean images; For the cross-attention module, we require that the cross-attention output calculated from the basic prompts and APV with adversarial examples be as consistent as possible. We calculate gradients with output loss and attention loss, respectively. Then we introduce the Local Random Timestep Gradient Ensemble strategy. We divide the entire timestep sequence into several groups and randomly select one timestep from each group for the gradient calculation mentioned above. Finally, we aggregate the gradients of all groups as the total gradient for updating adversarial examples in each iteration and execute alternating training and attack strategies similar to Anti-Dreambooth.

\subsection{Adversarial Prompt Vector}

Firstly, initializing from the instance prompt's CLIP encoding features, i.e. $P_{adv}=P_{new}$, we iteratively optimize the prompt vector by maximizing the loss between the UNet output and Gaussian noise following Eq. (\ref{eq2}) to obtain the APV. Therefore, we modify Eq. (\ref{eq4}) to Eq. (\ref{eq6}) as follows:
\begin{equation}\label{eq6}
    \begin{split}
    P_{adv} &\leftarrow \mathop{\arg\max}\limits_{P_{adv}}L(\theta,x_{0},P_{adv}) \\&\leftarrow \mathop{\max}\limits_{P_{adv}}\mathbb{E}_{x_{0},t,\epsilon}||\epsilon-\epsilon_{\theta}(x_{t+1},t,P_{adv})||_2^2.
    \end{split}
\end{equation}

This approach is inspired by \cite{mokady2023null} that updates prompt vectors to increase the generation quality. Since the APV does not need to be constrained during this process, and to alleviate the value gap between gradients at different timesteps, we normalize the gradient like \cite{dong2018boosting} and update the APV $P_{adv}$ using Eq. (\ref{eq7}), where $||g||=\frac{g}{||g||_1}$:
\begin{equation}
        \label{eq7}
        P_{adv}^{r'+1}=P_{adv}^{r'}+\eta ||\nabla_{P_{adv}}L(\theta,x_{0},P_{adv}^{r'})||.
\end{equation}

We randomly select $t \in \{0,1,..., 999\}$ to perform the attack for $r' \in R'$ iterations. Afterward, we complete the text-level attack and obtain $P_{adv}$ for subsequent attacks.

\begin{figure}[t]
    \centering
    \includegraphics[width=1.0\linewidth]{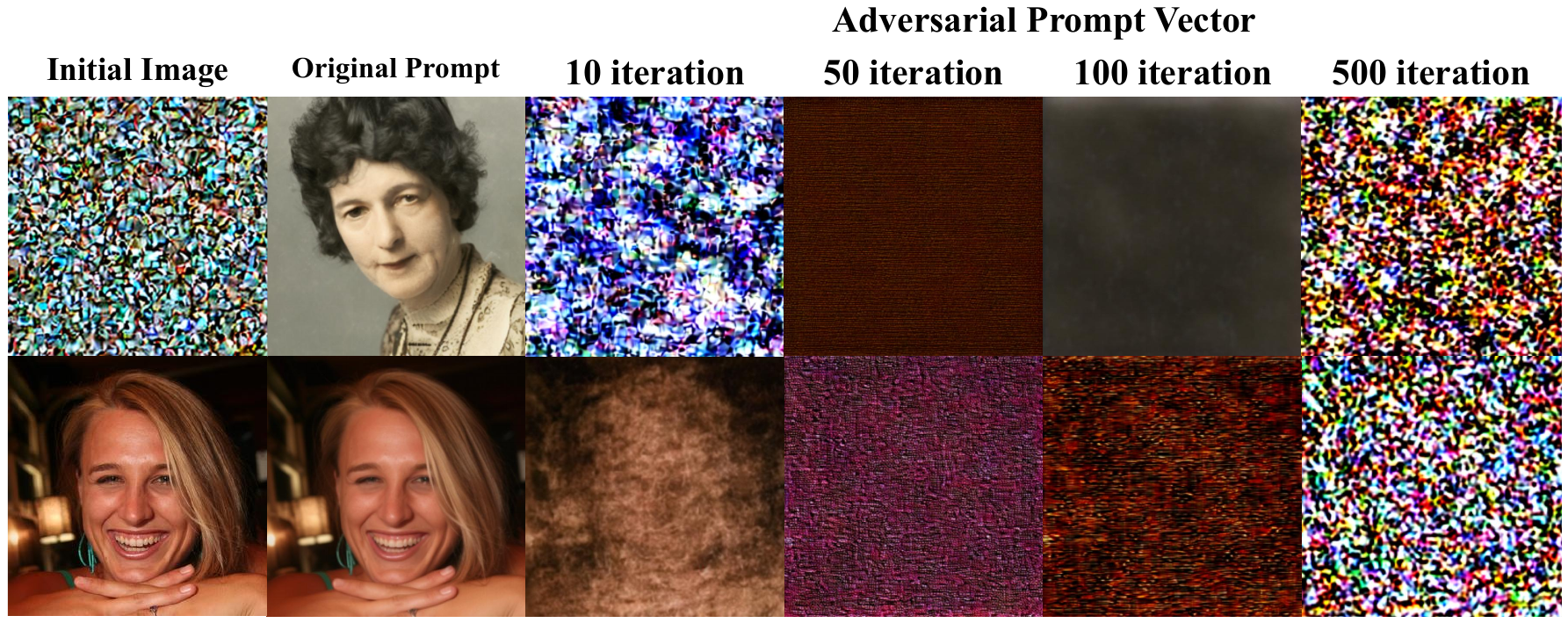}
    \caption{Generated images guided by original prompts (column 2) and APVs (column 3-6) starting from the prompt "A photo of a woman." Row 1: images from random noise. Row 2: images from DDIM-inverted initial image. APVs are created after 10, 50, 100, and 500 iterations.}
\label{fig3}
\end{figure}

To demonstrate the performance of APV, we generated images using random noise and noise from DDIM image inversion \cite{song2021denoising,dhariwal2021diffusion} as the start respectively, combined with the original prompt and APV at different iteration rounds. The results are shown in Fig. \ref{fig3}. It only takes 10 rounds of iterative attacks to make the diffusion model unable to generate meaningful images. With the increase of iteration rounds, the generated images contain less original structure and texture information of the original images and finally form a stable adversarial generation effect at 500 iteration rounds.
% When the iteration rounds of APV exceed 500, it fails to generate meaningful images using APV. As the rounds increase, the generated images contain less structure and texture information of the initial image, and a special pattern appears at 2000 rounds of iteration. 
This proves that the attack applied to the conditional vectors indeed possesses significant adversarial effects, and we can further enhance the attack effects of adversarial images by leveraging adversarial texts.

\subsection{Attention Module Disruption}
In the image-level adversarial attack, it has been proven effective in making the UNet model unable to predict the Gaussian noise added at the previous timestep during the diffusion process \cite{liang2023adversarial,van2023anti,wang2024simac,liu2024disrupting}. To combine the effect of APV with image-level adversarial examples and further disrupt pixel-level correlations in images, we propose to focus on disrupting attention modules of the UNet model.

\paragraph{Self-Attention Disruption}
The self-attention module calculates the correlation between image self-features, which can help the UNet capture the global dependency relationship between pixel-level positions, thereby improving the accuracy of generated images. To compromise the quality of image generation, we hope that the UNet fails to calculate the pixel correlation of adversarial examples. Assuming that the UNet model contains $S$ self-attention modules in total, the output of $s_{th}$ self-attention module is $f_s(\cdot)$. The self-attention loss can be calculated as Eq. (\ref{eq8}):

\vspace{-0.1cm}
\begin{equation}
    L_{SA}=\sum_{s=1}^{S}J(f_{s}(x_{0}+\delta_{adv}),f_{s}(x_{0})).
    \label{eq8}
\end{equation}

Since self-attention calculation does not need prompt vectors and the timestep is equal, we omit the prompt and timestep input in Eq. (\ref{eq8}) and use the same instance prompt for both the original image and adversarial examples. We use the Mean Square Error as $J(\cdot)$ to measure and maximize the distance between two results.

\vspace{-0.1cm}
\paragraph{Cross-Attention Disruption}

The cross-attention module calculates the attention weight between pixel features and prompt vectors, reflecting the degree of association between image pixels and the prompts. We calculate the cross-attention outputs of adversarial examples and instance prompts, as well as those of clean images and APV, and minimize the distance between them to achieve the integration of adversarial priors from APV with image-level adversarial examples. Assuming that the UNet model contains $C$ cross-attention modules in total, the output of $c_{th}$ cross-attention module is $f_c(\cdot)$. The cross-attention loss can be calculated as Eq. (\ref{eq9}). We also omit the timestep input in Eq. (\ref{eq9}), and choose Cosine Similarity distance as $J(\cdot)$ to measure and maximize the similarity between two results.

\vspace{-0.1cm}
\begin{equation}
    L_{CA}=\sum_{c=1}^{C}J(f_{c}(x_{0}+\delta_{adv},P_{new}),f_{c}(x_{0},P_{adv})).
    \label{eq9}
\end{equation}

After obtaining the $L_{SA}$ and $L_{CA}$, we use a hyper-parameter $\alpha_{1}$ to balance the two losses and obtain the attention loss $L_A$ in Eq. (\ref{eq10}). In Fig. \ref{attn}, we demonstrate the impact of DADiff on self- and cross-attention modules.

\begin{equation}
    L_{A}=\alpha_{1}L_{CA} + (1-\alpha_{1})L_{SA}.
    \label{eq10}
\end{equation}

\subsection{Local Random Timestep Gradient Ensemble}
Previous research has indicated that the timestep affects the effectiveness of attacking diffusion models. SimAC \cite{wang2024simac} selects timesteps with the highest gradient score for the attack, while DisDiff \cite{liu2024disrupting} allocates learning rates based on the Hybrid Quality Scores \cite{wang2023not} at different timesteps. However, they only compute gradients for a single timestep during a single iteration of PGD. 
% Randomly selecting the timestep can lead to fluctuations in loss and gradient score, thereby weakening the optimization effect. 
Although gradients at different timesteps vary in importance, they all contain information beneficial to optimizing adversarial examples. The gradient scores decrease with timesteps, with correlations between gradients at adjacent timesteps. Adding noise at multiple timesteps in the diffusion process is similar to noise augmentation in image classification. Integrating gradients from different timesteps can yield more robust adversarial gradients, improving the attack effect.

Therefore, we propose the Local Random Timestep Gradient Ensemble (LRTGE) strategy, combining the advantage of gradient ensemble and random gradient sampling. LRTGE divides the DDPM timesteps into $B$ equal segments and randomly selects one timestep within each segment to compute the gradient. The gradients from each segment are then summed and used as the update gradient for a single PGD attack. We formalize the process as Eq. (\ref{eq11}), where $||g||=\frac{g}{||g||_1}$. We calculate the gradients for $L_{cond}$ and $L_{A}$ separately because the analysis reveals a significant difference in the magnitude of the gradients obtained for the two. Since $L_{cond}$ is the core optimization function, we adopt the following strategy: first normalize two gradients separately, and then sum them up using a hyper-parameter $\alpha_{2}$.

% \vspace{-0.1cm}
\begin{equation}  
g^{r}=||\sum_{b=1}^{B}\nabla_{\delta}L_{cond}|| + \alpha_{2}||\sum_{b=1}^{B}\nabla_{\delta}L_{A}||,
\label{eq11}
\end{equation}  

After that, we use $g^{r}$ to update the image-level adversarial example in each PGD iteration (Eq. (\ref{eq12})).

\begin{equation}
        \delta_{adv}^{r+1}=clip_{\delta,\omega}(\delta_{adv}^{r}+\eta sign(g^{r})).
\label{eq12}
\end{equation}

\begin{figure}[tb]
    \centering
    \includegraphics[width=1.0\linewidth]{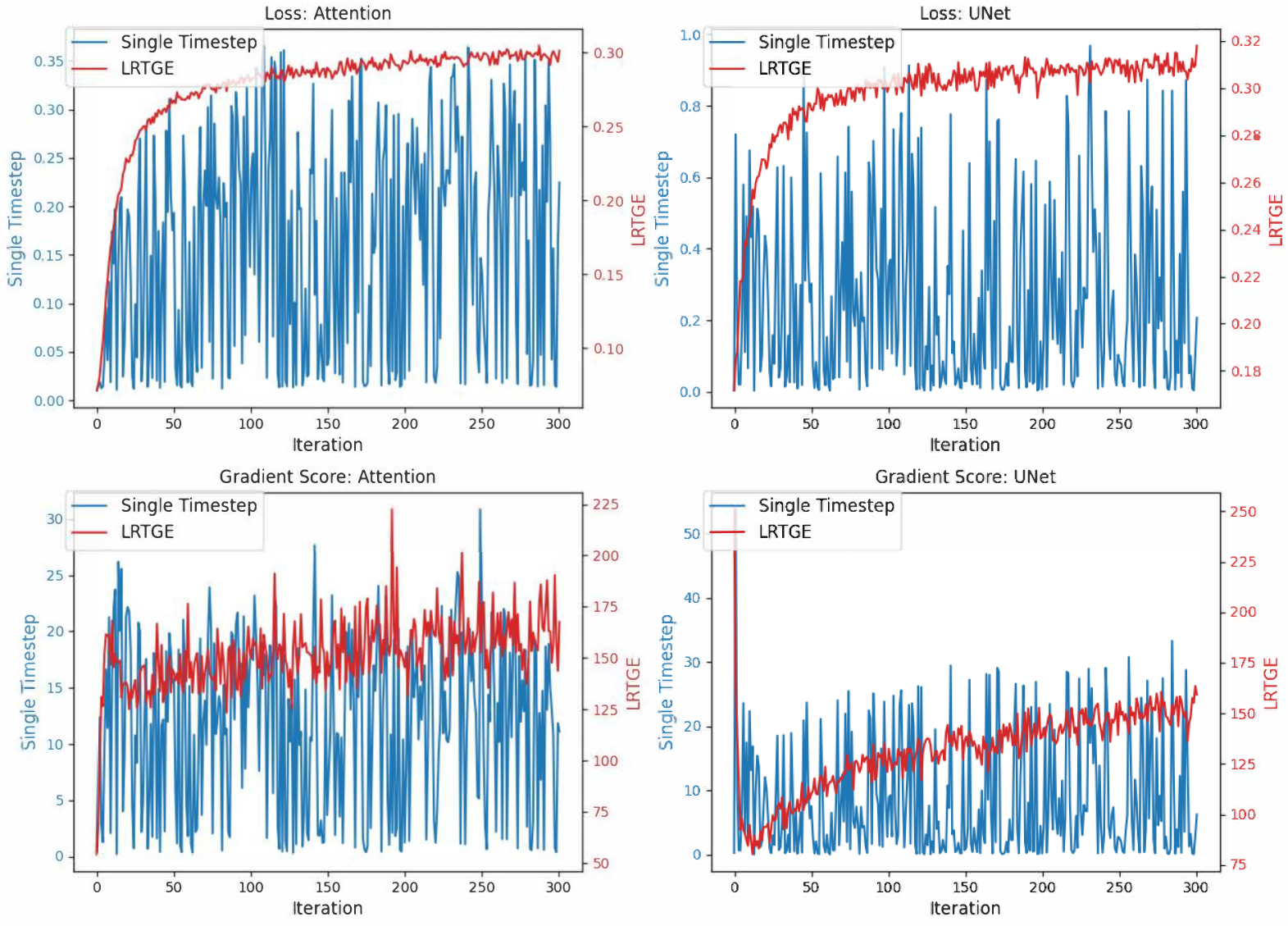}
    \caption{Comparisons of losses (first row) and gradient scores (second row) when performing PGD attack using single timestep (blue lines) and LRTGE (red lines). The total PGD iteration is 300 rounds (50 iterations ASPL \cite{van2023anti} with 6 iterations inside).}
\label{fig5}
\end{figure}

We implement DADiff using random single-timestep and LRTGE respectively, and draw on SimAC's gradient score calculation before normalization to observe the optimization process, as shown in Fig. \ref{fig5}. The results in the second row show that the gradient score of attention loss is larger than the UNet condition loss, so normalizing two gradients is necessary. The results also indicate that when using random single-timestep gradients, the loss exhibits significant instability, with large and irregular fluctuations in gradient scores, which is detrimental to the optimization of adversarial examples. In contrast, when optimization is performed using integrated multi-timestep gradients, the loss shows a steady upward trend, and the gradient scores remain generally stable, which is more beneficial for optimizing adversarial examples. Finally, the overall algorithm of the proposed DADiff is concluded in the Appendix.

\section{Experiments}
\label{sec:experiments}

\begin{table*}[htpb]
\centering
\caption{Quantitative evaluations.}
\label{table1}
\resizebox{\linewidth}{!}{
\begin{tabular}{ccccccc|ccccc}
\hline
\multirow{2}{*}{Dataset}   & \multirow{2}{*}{Method} & \multicolumn{5}{c|}{"a photo of sks person"}                                                 & \multicolumn{5}{c}{"a dslr portrait of sks person"}                                          \\ \cline{3-12} 
                           &                         & FDFR$\uparrow$ & ISM$\downarrow$ & SER-FQA$\downarrow$ & BRISQUE$\uparrow$ & FID$\uparrow$   & FDFR$\uparrow$ & ISM$\downarrow$ & SER-FQA$\downarrow$ & BRISQUE$\uparrow$ & FID$\uparrow$   \\ \hline
\multirow{6}{*}{VGGFace2}  & w/o Protect             & 0.00           & 0.72            & 0.77                & 10.96             & 185.61          & 0.00           & 0.53            & 0.65                & 1.27              & 218.03          \\ \cline{2-12} 
                           & AdvDM                   & 0.33           & 0.17            & 0.46                & 15.33             & 341.05          & 0.00           & 0.23            & 0.54                & 15.15             & 274.12          \\
                           & Anti-DB                 & 0.78           & 0.10            & 0.19                & 40.06             & 373.48          & 0.10           & 0.22            & 0.46                & 21.18             & 329.76          \\
                           & SimAC                   & 0.72           & 0.15            & 0.17                & 34.88             & 399.45          & 0.22           & 0.38            & 0.51                & 20.46             & 266.95          \\
                           & DisDiff                 & 0.97           & 0.02            & 0.01                & 50.45             & 427.12          & 0.20           & 0.30            & 0.55                & 20.92             & 281.64          \\
                           & DADiff(Ours)            & \textbf{0.99}  & \textbf{0.00}   & \textbf{0.00}       & \textbf{53.50}    & \textbf{526.90} & \textbf{0.91}  & \textbf{0.04}   & \textbf{0.08}       & \textbf{43.79}    & \textbf{462.63} \\ \hline
\multirow{6}{*}{CelebA-HQ} & w/o Protect             & 0.00           & 0.77            & 0.84                & 27.35             & 139.33          & 0.07           & 0.49            & 0.79                & 2.79              & 226.92          \\ \cline{2-12} 
                           & AdvDM                   & 0.67           & 0.06            & 0.53                & 14.41             & 294.45          & 0.10           & 0.06            & 0.68                & 16.40             & 258.36          \\
                           & Anti-DB                 & 0.77           & 0.04            & 0.17                & 54.97             & 336.57          & 0.17           & 0.12            & 0.70                & 17.08             & 291.99          \\
                           & SimAC                   & 0.97           & 0.03            & 0.03                & \textbf{60.15}             & 478.24 & 0.47           & 0.16            & 0.55                & 19.50             & 228.09          \\
                           & DisDiff                 & 0.98           & 0.03            & 0.01                & 58.15    & 471.06          & 0.56           & 0.12            & 0.54                & 20.92             & 356.36          \\
                           & DADiff(Ours)            & \textbf{0.98}  & \textbf{0.01}   & \textbf{0.00}       & 59.77             & \textbf{479.66}          & \textbf{0.88}  & \textbf{0.03}   & \textbf{0.15}       & \textbf{36.41}    & \textbf{383.60} \\ \hline
\end{tabular}
}
\end{table*}

\subsection{Experimental Setups}
\paragraph{Datasets and Comparisons}
We evaluate DADiff on two widely-used facial datasets: CelebA-HQ \cite{liu2015deep} and VGGFace2 \cite{cao2018vggface2}. We randomly choose 16 identities in each dataset and 8 different images are prepared for each identity. The images are equally split into the clean set and the perturbed set. All images are center-cropped and resized to $512 \times 512$. We compare DADiff with most related AdvDM \cite{liang2023adversarial}, Anti-DB \cite{van2023anti}, SimAC \cite{wang2024simac} and DisDiff \cite{liu2024disrupting}.
\paragraph{Evaluation Metrics}
To gauge the effectiveness of DADiff, we employ several metrics to evaluate the quality of images produced by the disrupted Dreambooth models. We assess the Face Detection Failure Rate (FDFR) using the RetinaFace detector \cite{deng2020retinaface}. For successfully detected faces, we further leverage the ArcFace recognizer \cite{deng2019arcface} to obtain face recognition embeddings and compute the cosine distance to the mean face embedding of the user's entire set of clean images, a process called Identity Score Matching (ISM). Additionally, use SER-FQA \cite{terhorst2020ser}, BRISQUE \cite{mittal2012no}, and FID \cite{heusel2017gans} to further evaluate the generated image quality. The best results are emphasized in \textbf{bold}.

\vspace{-0.1cm}
\paragraph{Models and Parameters}
We use two versions of pre-trained open-source SD models from HuggingFace \cite{von_platen_etal_2022_diffusers} in the main text: v1.5 and v2.1, and use v1.4, v1.5 and v2.1 in the Appendix for further comparisons. 
For training Dreambooth, the batch size is set to 2 and the learning rate is set to $5 \times 10^{-7}$ for 1000 training steps. 
Unless otherwise specified, we use SD v1.5 as the surrogate model. \textbf{All adversarial attacks, including APV, are generated with "\textit{A photo of sks person}" as the original white-box instance prompt.} The learning rates $\eta$ of the adversarial examples in DADiff are all set to 0.005. In APV generation, the number of iterations $R'=500$. APV does not need to be constrained, but the image-level perturbation constraint $\omega=0.05$. In the image-level attack, we disrupt all self- and cross-attention layers in the upsampling module of the UNet model. We equally divide the overall timesteps $T=1000$ into $B=25$ segments to calculate the ensemble gradients. The hyper-parameters are set to $\alpha_{1}=0.5$ and $\alpha_{2}=0.4$. We follow the same settings of ASPL \cite{van2023anti} to perform $R=300$ rounds of PGD attack in total. We display cross-prompt, keyword mismatch, cross-model and cross-mechanism comparisons in the main text. Detailed ablations, additional experimental results, and more visualizations are provided in the Appendix.

% We use "\textit{A dslr portrait of sks person}" as the black-box prompt and "\textit{asdf}" as the black-box keyword in the main text. Furthermore, we use "\textit{A photo of sks person looking at the mirror}", "\textit{A photo of sks person in front of Eiffel Tower}" as black-box inference prompts to evaluate the effectiveness in the Appendix.

\subsection{Comparisons and Evaluations}

\begin{table}[tb]
\centering
\caption{Keyword mismatch evaluations.}
\label{table-cross}
\resizebox{\linewidth}{!}{
\begin{tabular}{ccccccc}
\hline
\multirow{2}{*}{\begin{tabular}[c]{@{}c@{}}Dreambooth\\ Prompt\end{tabular}}         & \multirow{2}{*}{Method} & \multicolumn{5}{c}{"a photo of asdf person"}                                     \\ \cline{3-7} 
                                                                                     &                         & FDFR↑         & ISM↓          & SER-FQA↓      & BRISQUE↑       & FID↑            \\ \hline
\multirow{11}{*}{\begin{tabular}[c]{@{}c@{}}"A photo of\\ asdf person"\end{tabular}} & AdvDM                   & 0.08          & 0.18          & 0.53          & 13.25          & 339.24          \\
                                                                                     & Anti-DB                 & 0.47          & 0.14          & 0.35          & 23.32          & 347.25          \\
                                                                                     & SimAC                   & 0.59          & 0.04          & 0.35          & 29.02          & 348.97          \\
                                                                                     & DisDiff                 & 0.88          & 0.02          & 0.10          & 37.53          & 417.05          \\
                                                                                     & DADiff(Ours)            & \textbf{0.91} & \textbf{0.02} & \textbf{0.08} & \textbf{44.58} & \textbf{443.06} \\ \cline{2-7} 
                                                                                     &                         & \multicolumn{5}{c}{"a dslr portrait of asdf person"}                             \\ \cline{2-7} 
                                                                                     & AdvDM                   & 0.11          & 0.15          & 0.60          & 15.75          & 284.37          \\
                                                                                     & Anti-DB                 & 0.11          & 0.17          & 0.60          & 12.68          & 277.85          \\
                                                                                     & SimAC                   & 0.08          & 0.17          & 0.70          & 5.59           & 227.91          \\
                                                                                     & DisDiff                 & 0.13          & 0.23          & 0.60          & 19.32          & 253.16          \\
                                                                                     & DADiff(Ours)            & \textbf{0.25} & \textbf{0.09} & \textbf{0.14} & \textbf{24.80} & \textbf{289.75} \\ \hline
\end{tabular}
}
\end{table}

\paragraph{Quantitative Results}
We compared our method with mainstream approaches on SD v1.5, and the results are presented in Table \ref{table1}. On the white-box instance prompt ("\textit{a photo of a sks person}"), compared to the baseline AdvDM and Anti-DB, state-of-the-art methods yield better performance. SimAC yields a competitive BRISQUE value on the CelebA-HQ dataset, but its performance on the VGGFace2 dataset experiences a significant decline. DADiff produces poorer image quality on both datasets, not only struggling to generate recognizable faces (higher FDFR and lower ISM) but also worsening the invisibility of the generated images (lower SER-FQA, higher BRISQUE, and FID). 

On the black-box inference prompt (\textit{"a dslr portrait of sks person"}), DADiff demonstrates significant advantages on both datasets, with an increase of over 32\% in the proportion of images where faces cannot be generated (FDFR), and a substantial decline of over 30\% in the quality of the generated images. In the Appendix, we further evaluate the generation results of several methods on more black-box inference prompts, all indicating DADiff's remarkable advantage in cross-prompt transferability. This suggests that DADiff more thoroughly disrupts the correlation between the keyword (\textit{"sks"}) and the target person's image.

\vspace{-0.1cm}
\paragraph{Qualitative Results}

\begin{figure}[tb]
    \centering
    \includegraphics[width=\linewidth]{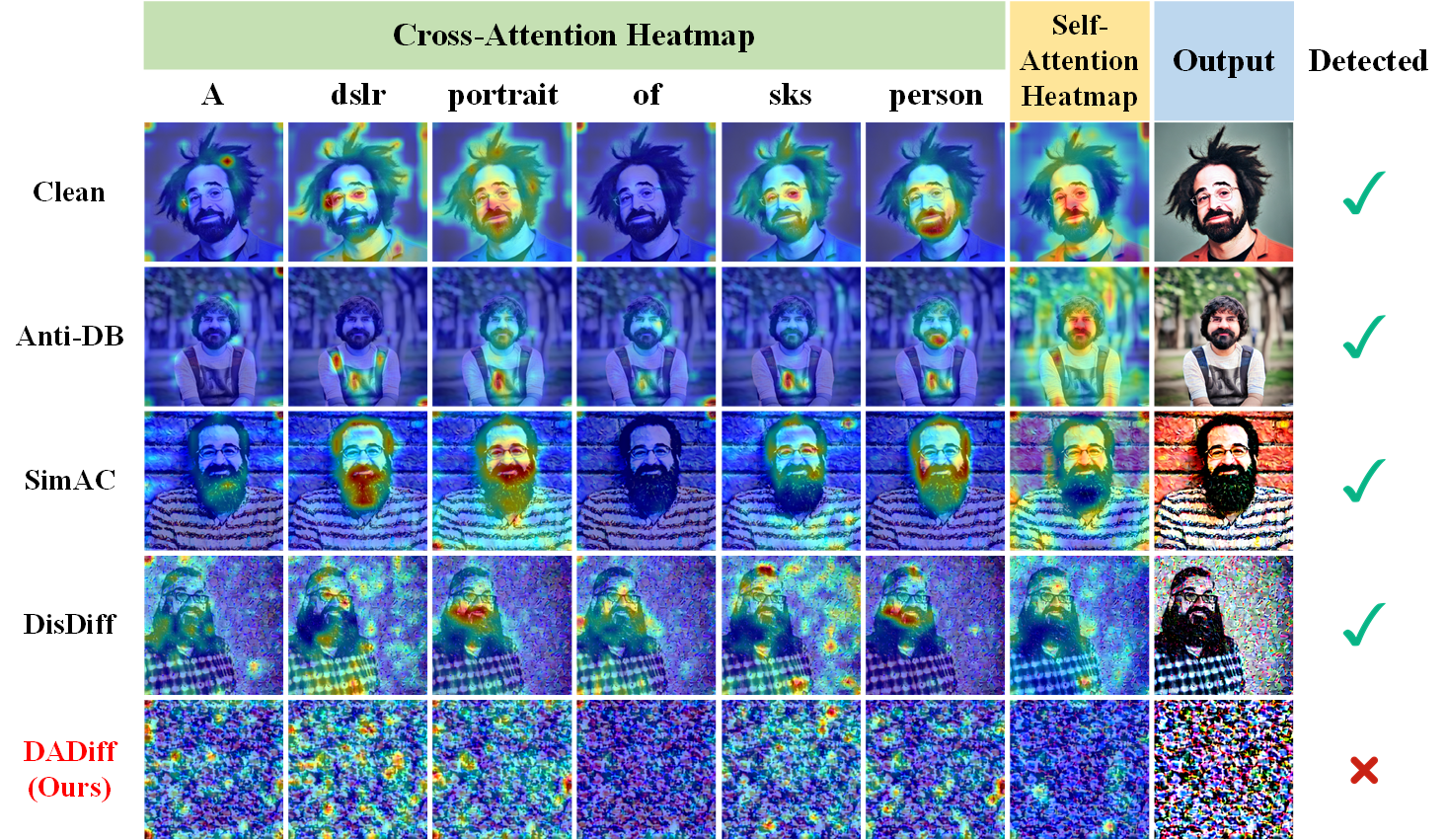}
    \caption{Attention heatmaps of generated images from vanilla Dreambooth training and different anti-customization methods.}
\label{attn}
\end{figure}

Fig. \ref{fig1} presents the generation quality of major comparisons and DADiff under different prompts. It shows that DADiff causes more thorough disruption to the generation quality, making the generated images more unrecognizable, and exhibiting better transferability among different black-box inference prompts. Visualization results on more IDs are shown in Section D of the Appendix.

To further analyze the impact of adversarial examples on the internal modules of the UNet model, Fig. \ref{attn} shows the generated image's cross- and self-attention heatmaps on the UNet model. The first row presents the heatmap of the generated image from the vanilla Dreambooth-trained model, and the rest are heatmaps of the generated image on the adversarial Dreambooth-trained model. We select the $16\times 16$-sized cross- and self-attention module to draw the heatmap and display it at timestep $t=500$. The self-attention heatmap is drawn using the mean attention of all pixel tokens, reflecting the high-attention areas of the image globally. The figure shows that the Dreambooth model trained with adversarial examples generated by DADiff is difficult to associate between prompt keywords and images and is unable to effectively capture the global correlation of pixels. The attack on the self-attention module undermines the model's ability to learn the correlation between image pixels, while the guidance of APV in the cross-attention module also exacerbates the damage to the model's ability to learn the correlation between the prompt words and the target person's images. This reflects the effectiveness of attacks on self-attention and cross-attention modules. 

\begin{table}[tb]
\centering
\caption{Cross-model evaluations.}
\label{table3}
\resizebox{\linewidth}{!}{
\begin{tabular}{ccccccc}
\hline
\multirow{2}{*}{Surrogate}            & \multirow{2}{*}{Method} & \multicolumn{5}{c}{SD v1.5}                                                      \\ \cline{3-7} 
                                      &                         & FDFR↑         & ISM↓          & SER-FQA↓      & BRISQUE↑       & FID↑            \\ \hline
                                      & w/o Protect             & 0.07          & 0.55          & 0.62          & 11.97          & 232.98          \\ \hline
\multirow{3}{*}{v1.5}                 & Anti-DB                 & 0.45          & 0.14          & 0.28          & 31.12          & 351.62          \\
                                      & DisDiff                 & 0.75          & 0.10          & 0.20          & 38.55          & 398.72          \\
                                      & DADiff(Ours)            & \textbf{0.97} & \textbf{0.06} & \textbf{0.04} & \textbf{43.60} & \textbf{444.81} \\ \hline
\multirow{3}{*}{v2.1}                 & Anti-DB                 & 0.50          & 0.18          & 0.31          & 32.87          & 338.39          \\
                                      & DisDiff                 & 0.48          & 0.17          & 0.31          & 31.95          & 354.58          \\
                                      & DADiff(Ours)            & \textbf{0.63} & \textbf{0.12} & \textbf{0.23} & \textbf{34.18} & \textbf{359.75} \\ \hline
\multicolumn{1}{l}{\multirow{2}{*}{}} & \multicolumn{1}{l}{}    & \multicolumn{5}{c}{SD v2.1}                                                      \\ \cline{2-7} 
\multicolumn{1}{l}{}                  & w/o Protect             & 0.06          & 0.41          & 0.71          & 10.04          & 207.54          \\ \hline
\multirow{3}{*}{v1.5}                 & Anti-DB                 & 0.09          & 0.32          & 0.58          & 17.99          & 225.15          \\
                                      & DisDiff                 & 0.78          & 0.05          & 0.30          & 32.71          & 294.28          \\
                                      & DADiff(Ours)            & \textbf{0.81} & \textbf{0.04} & \textbf{0.24} & \textbf{36.88} & \textbf{311.12} \\ \hline
\multirow{3}{*}{v2.1}                 & Anti-DB                 & 0.19          & 0.30          & 0.31          & 29.55          & 277.85          \\
                                      & DisDiff                 & 0.64          & 0.10          & 0.30          & 35.52          & 339.83          \\
                                      & DADiff(Ours)            & \textbf{0.72} & \textbf{0.07} & \textbf{0.24} & \textbf{38.73} & \textbf{366.92} \\ \hline
\end{tabular}
}
\end{table}

\vspace{-0.1cm}
\paragraph{Keyword Mismatch Evaluations}
Image protectors often have difficulty predicting what keywords [A] the adversary will use to refer to the target individual. To this end, we evaluate the generation quality after using mismatched keywords for Dreambooth training. We choose "\textit{sks}" as the keyword for generating adversarial examples, and use "\textit{asdf}" to train and evaluate the quality of Dreambooth's generation. As shown in Table \ref{table-cross}, DADiff still exhibits the best performance in the case of keyword mismatch. In the black-box inference prompt, DADiff is still better than that of existing methods. The interference with the self-attention module may weaken the dependence of adversarial examples on specific prompt words, thereby achieving stronger attack effects in the case of keyword mismatch.

\vspace{-0.1cm}
\paragraph{Cross-Model Evaluations}
We utilize the SD v1.5 and v2.1 models, respectively, employing "\textit{a photo of sks person}" as the instance prompt to generate adversarial examples. Subsequently, we use the VGGFace2 dataset, and "\textit{a photo of sks person}", "\textit{a dslr portrait of sks person}" as inference prompts to generate images and calculate the average results of these two inference prompts. We only compare DADiff with the baseline Anti-DB and the most recent DisDiff in the main text. Table \ref{table3} shows that, whether the adversarial examples generated by the v1.5 model are used for Dreambooth training of the v2.1 model or vice versa, DADiff exhibits the strongest disruption. This indicates that DADiff has more thoroughly disrupted the UNet model's association between the keywords \textit{sks} and the ID image. We include demonstrations of the adversarial examples' performance on the v1.4 model in Table 2 of the Appendix.

\vspace{-0.1cm}
\paragraph{Cross-Mechanism Evaluations}
We also evaluate the transferability of adversarial examples under different fine-tuning diffusion mechanisms. As shown in Table \ref{table-mechanism}, we generate adversarial examples using the Dreambooth mechanism and then apply them for fine-tuning LoRA and Texture Inversion (TI). It indicates that, despite significant differences in fine-tuning principles, DADiff still achieves the best performance in adversarial attacks across mechanisms. This may be because the model structures employed by different fine-tuning mechanisms remain similar, and the destructive impact on the model's internal components is preserved across various generation mechanisms. The Appendix demonstrates the generated images of different mechanisms under adversarial examples. DADiff exhibits significant advantages both quantitatively and qualitatively.

\begin{table}[tb]
\centering
\caption{Cross-mechanism evaluations.}
\label{table-mechanism}
\resizebox{\linewidth}{!}{
\begin{tabular}{ccccccc}
\hline
\multirow{2}{*}{Mechanism} & \multirow{2}{*}{Method} & \multirow{2}{*}{FDFR↑} & \multirow{2}{*}{ISM↓} & \multirow{2}{*}{SER-FQA↓} & \multirow{2}{*}{BRISQUE↑} & \multirow{2}{*}{FID↑} \\
                           &                         &                        &                       &                           &                           &                       \\ \hline
\multirow{4}{*}{LoRA}      & w/o Protect             & 0.06                   & 0.58                  & 0.71                      & 11.90                     & 178.69                \\ \cline{2-7} 
                           & Anti-DB                 & 0.00                   & 0.53                  & 0.69                      & 17.57                     & 220.51                \\
                           & DisDiff                 & 0.28                   & 0.32                  & 0.51                      & 23.21                     & 247.18                \\
                           & DADiff(Ours)            & \textbf{0.31}          & \textbf{0.29}         & \textbf{0.42}             & \textbf{30.31}            & \textbf{285.24}       \\ \hline
\multirow{4}{*}{TI}        & w/o Protect             & 0.06                   & 0.38                  & 0.44                      & 2.73                      & 233.36                \\ \cline{2-7} 
                           & Anti-DB                 & 0.19                   & 0.14                  & 0.29                      & 13.34                     & 348.68                \\
                           & DisDiff                 & 0.25                   & 0.10                  & 0.46                      & 14.31                     & 370.03                \\
                           & DADiff(Ours)            & \textbf{0.38}          & \textbf{0.05}         & \textbf{0.17}             & \textbf{28.80}            & \textbf{382.64}       \\ \hline
\end{tabular}
}
\end{table}

\section{Conclusion}
\label{sec:conclusion}

We propose DADiff, a diffusion model anti-customization method based on two-stage adversarial attacks. We make full use of all the inputs and key components of the diffusion model during the attack and achieve better results than existing methods in qualitative and quantitative evaluations across multiple metrics. Our research also confirms the idea that for models with complex inputs and complex structures, each input and module may have potential adversarial risks. This insight will guide us to continuously optimize and improve the effectiveness and practicality of adversarial examples of diffusion models in more practical application scenarios such as social platforms and complex fine-tuning algorithms in our future work.

\clearpage

{
    \small
    \bibliographystyle{ieeenat_fullname}
    \bibliography{reference}
}

\clearpage

\appendix

%%%%%%%%% BODY TEXT - ENTER YOUR RESPONSE BELOW
\section{Additional Declarations and Algorithm}

In the quantitative experiments in the main text and appendix, unless otherwise stated, the values of different metrics are average values of the generated quality of selected images on the VGGFace2 and CelebA-HQ datasets, and all experiments without additional declarations use the SD v1.5 model to generate adversarial examples. We demonstrate the overall pipeline of DADiff in Algorithm \ref{ag1}.

\begin{algorithm}[h]
\caption{Dual Anti-Diffusion (DADiff)}
\label{ag1}
\textbf{Input}: Stable Diffusion model $SD$. Original images $X$. Instance prompt $P$. Iterations for APV $R'$. Iterations for image-level perturbation $R$ containing outside rounds $t_{out}$ and inside rounds $t_{in}$. Iterations for Dreambooth training $t_1$. Hyperparameters for Dreambooth training $\alpha_{DB}$, attention loss fusion $\alpha_{1}$, and gradient fusion $\alpha_{2}$. Overall diffusion timesteps $T$ and number of segments $B$.\\
\textbf{Output}: Protected images $X_{adv}$.
\begin{algorithmic}[1]
\Procedure{DB}{$SD, X, P, t_{1}$}
\For{$t'$ in $t_1$}
\State Random Sample timestep $t$ from $T$,\;
\State Calculate $L_{db}$ using Eq. (3),\;
\State $SD \leftarrow SD - \alpha_{DB} \nabla L_{db}$,\;
\EndFor
\State \Return $SD$.\;
\EndProcedure
\Procedure{APV}{$SD, X, P, R'$}
\For{$r'$ in $R'$}
\State Random Sample timestep $t$ from $T$,\;
\State Calculate $L_{cond}$ using Eq. (2),\;
\State Update $P_{adv}$ using Eq. (7),\;
\EndFor
\State \Return $P_{adv}$.\;
\EndProcedure
\State
\State // \textbf{Generating APV}
\State $P_{adv}$ $\leftarrow$ APV($SD, X, P, R'$),\;
\State Initialize $\delta$,\;

\For{$r$ in $t_{out}$}

\State // \textbf{Dreambooth Training}
\State $SD$ $\leftarrow$ \Call{$DB$}{$SD, X, P, t_1$},\;

\For{$t''$ in $t_{in}$}

\State // \textbf{LRTGE}
\State Divide timesteps $T$ into $B$ segments,\;
\For{$b$ in $B$}
\State Random Sample timestep $t_b$ from $b$,\;
\State Calculate $L_{cond}$ using Eq. (2),\;
\State $g_{b}^{cond}=\nabla_{\delta}L_{cond}$,\;

\State // \textbf{Attention Disruption}
\State Calculate $L_{SA}$ using Eq. (8),\;
\State Calculate $L_{CA}$ with $P_{adv}$ using Eq. (9),\;
\State $L_{A}=\alpha_{1}L_{SA}+(1-\alpha_{1})L_{CA}$,\;
\State $g_{b}^{A}=\nabla_{\delta}L_{A}$,\;
\EndFor
\State $g_{cond}=\sum_{b=1}^{B}g_{b}^{cond}$,\;
\State $g_{A}=\sum_{b=1}^{B}g_{b}^{A}$,\;
\State $g^{r}=||g_{cond}||+\alpha_{2}||g_{A}||$ using Eq. (11),\;
\State Update $\delta$ using Eq. (12),\;
\EndFor
\State $X_{adv}=X+\delta$,\;
\State // \textbf{Dreambooth Training}
\State $SD \leftarrow $ \Call{$DB$}{$SD, X_{adv}, P, t_1$}.\;
\EndFor
\State \Return $X_{adv}$.\;

\end{algorithmic}
\end{algorithm}

\section{Results on More Inference Prompts}

Due to space constraints, we only present the quantitative results on the white-box instance prompt "\textit{a photo of sks person}" and a black-box inference prompt "\textit{a dslr portrait of sks person}" in Table 1 of the main text. To further demonstrate the effectiveness of DADiff in cross-prompt transferability, we generate adversarial examples using the white-box instance prompt "\textit{a photo of sks person}", and present quantitative comparisons in Table \ref{fulutable-prompt} for the other three black-box inference prompts: "\textit{a photo of sks person looking at the mirror}", "\textit{a photo of sks person in front of Eiffel Tower}", and "\textit{a photo of sks person sitting on the chair}". DADiff achieves the best results in all metrics on all black-box inference prompts.

\begin{table*}[htpb]
\centering
\caption{More cross-prompt evaluations.}
\label{fulutable-prompt}
\resizebox{\linewidth}{!}{
\begin{tabular}{ccccccc|ccccc|ccccc}
\hline
\multirow{2}{*}{Dataset}   & \multirow{2}{*}{Method} & \multicolumn{5}{c|}{"a photo of sks person looking at the mirror"}               & \multicolumn{5}{c|}{"a photo of sks person in front of Eiffel tower"}            & \multicolumn{5}{c}{"a photo of sks person sitting on the chair"}                 \\ \cline{3-17} 
                           &                         & FDFR↑         & ISM↓          & SER-FQA↓      & BRISQUE↑       & FID↑            & FDFR↑         & ISM↓          & SER-FQA↓      & BRISQUE↑       & FID↑            & FDFR↑         & ISM↓          & SER-FQA↓      & BRISQUE↑       & FID↑            \\ \hline
\multirow{6}{*}{VGGFace2}  & w/o Protect             & 0.03          & 0.41          & 0.45          & 17.43          & 295.31          & 0.13          & 0.10          & 0.27          & 12.67          & 382.11          & 0.20          & 0.16          & 0.35          & 17.54          & 355.80          \\ \cline{2-17} 
                           & AdvDM                   & 0.17          & 0.18          & 0.27          & 30.19          & 403.46          & 0.67          & 0.03          & 0.20          & 10.39          & 403.25          & 0.00          & 0.25          & 0.32          & 22.89          & 407.38          \\
                           & Anti-DB                 & 0.30          & 0.16          & 0.38          & 22.76          & 364.13          & 0.20          & 0.07          & 0.19          & 14.61          & 427.00          & 0.43          & 0.08          & 0.24          & 22.83          & 393.91          \\
                           & SimAC                   & 0.25          & 0.28          & 0.24          & 20.07          & 364.47          & 0.17          & 0.07          & 0.25          & 23.94          & 423.26          & 0.25          & 0.15          & 0.16          & 21.31          & 386.93          \\
                           & DisDiff                 & 0.50          & 0.14          & 0.25          & 18.98          & 384.13          & 0.25          & 0.06          & 0.31          & 11.04          & 438.62          & 0.81          & 0.03          & 0.08          & 26.18          & 448.35          \\
                           & DADiff(Ours)            & \textbf{1.00} & \textbf{0.00} & \textbf{0.06} & \textbf{38.35} & \textbf{527.46} & \textbf{0.88} & \textbf{0.01} & \textbf{0.08} & \textbf{34.65} & \textbf{466.10} & \textbf{1.00} & \textbf{0.00} & \textbf{0.00} & \textbf{41.01} & \textbf{525.60} \\ \hline
\multirow{6}{*}{CelebA-HQ} & w/o Protect             & 0.07          & 0.46          & 0.69          & 24.03          & 194.30          & 0.23          & 0.07          & 0.44          & 19.33          & 412.04          & 0.23          & 0.16          & 0.58          & 21.64          & 364.87          \\ \cline{2-17} 
                           & AdvDM                   & 0.10          & 0.13          & 0.52          & 22.97          & 372.62          & 0.67          & 0.01          & 0.21          & 8.49           & 403.24          & 0.13          & 0.06          & 0.40          & 21.50          & 398.33          \\
                           & Anti-DB                 & 0.80          & 0.03          & 0.22          & 29.70          & 419.62          & 0.40          & 0.01          & 0.20          & 15.05          & 416.66          & 0.96          & 0.02          & 0.03          & 21.42          & 459.50          \\
                           & SimAC                   & 0.93          & 0.01          & 0.01          & 32.23          & 500.81          & 0.53          & 0.01          & 0.03          & 18.88          & 436.13          & 0.99          & 0.01          & 0.01          & 28.40          & 465.87          \\
                           & DisDiff                 & 1.00          & 0.01          & 0.00          & 32.14          & 501.42          & 0.69          & 0.01          & 0.15          & 24.26          & 396.48          & 0.99          & 0.01          & 0.01          & 31.68          & 439.96          \\
                           & DADiff(Ours)            & \textbf{1.00} & \textbf{0.00} & \textbf{0.00} & \textbf{38.74} & \textbf{516.21} & \textbf{0.88} & \textbf{0.00} & \textbf{0.01} & \textbf{33.22} & \textbf{526.97} & \textbf{1.00} & \textbf{0.00} & \textbf{0.00} & \textbf{35.75} & \textbf{488.70} \\ \hline
\end{tabular}
}
\end{table*}

\section{More Cross-Model Evaluations}

In the main text, we perform a cross-model analysis of the anti-customization effects of SD v1.5 and SD v2.1 models. Here, we add SD v1.4 and utilize the average value of image generation quality guided by five inference prompts for a more comprehensive evaluation. The experimental results are presented in Table \ref{fulutable-model}. The results indicate that DADiff demonstrates the best anti-customization effect on each model, particularly in cross-model scenarios, showcasing significant advantages. Compared to existing methods, DADiff enhances the average value of cross-model transferability by 2\%-40\%. Considering that each value signifies the average quality evaluation on several generated images of 32 IDs across five prompts on two datasets, it suggests that DADiff possesses broader transferability across models, prompts, and facial datasets. This further proves that a more comprehensive and thorough utilization of key components and inputs of the diffusion model in adversarial attacks can achieve more effective anti-customizations.

\begin{table*}[hptb]
\centering
\caption{Cross-model evaluations.}
\label{fulutable-model}
\resizebox{\linewidth}{!}{
\begin{tabular}{ccccccc|ccccc|ccccc}
\hline
\multirow{2}{*}{Surrogate} & \multirow{2}{*}{Method} & \multicolumn{5}{c|}{SD v1.5}                                                     & \multicolumn{5}{c|}{SD v1.4}                                                     & \multicolumn{5}{c}{SD v2.1}                                                      \\ \cline{3-17} 
                           &                         & FDFR↑         & ISM↓          & SER-FQA↓      & BRISQUE↑       & FID↑            & FDFR↑         & ISM↓          & SER-FQA↓      & BRISQUE↑       & FID↑            & FDFR↑         & ISM↓          & SER-FQA↓      & BRISQUE↑       & FID↑            \\ \hline
\multicolumn{2}{c}{w/o Protect}                      & 0.07          & 0.55          & 0.62          & 11.97          & 232.98          & 0.09          & 0.39          & 0.58          & 15.73          & 310.72          & 0.06          & 0.41          & 0.71          & 10.04          & 207.54          \\ \hline
\multirow{4}{*}{v1.5}      & Anti-DB                 & 0.37          & 0.15          & 0.27          & 24.49          & 377.66          & 0.39          & 0.12          & 0.29          & 21.10          & 391.21          & 0.09          & 0.32          & 0.58          & 17.99          & 225.15          \\
                           & SimAC                   & 0.26          & 0.25          & 0.29          & 20.55          & 356.03          & 0.04          & 0.37          & 0.50          & 18.22          & 343.94          & 0.79          & 0.04          & 0.23          & 30.45          & 284.75          \\
                           & DisDiff                 & 0.62          & 0.13          & 0.25          & 25.45          & 399.73          & 0.31          & 0.22          & 0.36          & 22.75          & 374.05          & 0.78          & 0.05          & 0.30          & 32.71          & 294.28          \\
                           & DADiff(Ours)            & \textbf{0.93} & \textbf{0.06} & \textbf{0.03} & \textbf{40.59} & \textbf{464.56} & \textbf{0.93} & \textbf{0.06} & \textbf{0.10} & \textbf{42.21} & \textbf{444.20} & \textbf{0.81} & \textbf{0.03} & \textbf{0.22} & \textbf{36.88} & \textbf{311.12} \\ \hline
\multirow{4}{*}{v1.4}      & Anti-DB                 & 0.50          & 0.10          & 0.26          & 26.00          & 396.45          & 0.48          & 0.09          & 0.32          & 22.18          & 397.46          & 0.16          & 0.24          & 0.36          & 34.74          & 340.39          \\
                           & SimAC                   & 0.19          & 0.31          & 0.43          & 18.52          & 344.69          & 0.29          & 0.20          & 0.36          & 28.90          & 395.83          & 0.59          & 0.12          & 0.13          & 42.60          & 407.08          \\
                           & DisDiff                 & 0.61          & 0.10          & 0.16          & 31.20          & 409.17          & 0.19          & 0.23          & 0.31          & 22.22          & 344.71          & 0.41          & 0.14          & 0.28          & \textbf{43.33} & 389.85          \\
                           & DADiff(Ours)            & \textbf{0.86} & \textbf{0.08} & \textbf{0.12} & \textbf{33.25} & \textbf{441.07} & \textbf{0.94} & \textbf{0.04} & \textbf{0.04} & \textbf{38.72} & \textbf{478.57} & \textbf{0.64} & \textbf{0.11} & \textbf{0.13} & 40.42          & \textbf{414.60} \\ \hline
\multirow{4}{*}{v2.1}      & Anti-DB                 & 0.10          & 0.26          & 0.42          & 20.68          & 349.72          & 0.25          & 0.20          & 0.33          & 21.54          & 368.05          & 0.16          & 0.22          & 0.45          & 25.63          & 315.55          \\
                           & SimAC                   & 0.44          & 0.18          & 0.33          & 29.00          & 394.69          & 0.24          & 0.29          & 0.41          & 24.23          & 374.57          & 0.70          & 0.14          & 0.17          & 43.02          & 389.57          \\
                           & DisDiff                 & 0.39          & 0.13          & 0.32          & 26.27          & 397.12          & 0.46          & 0.12          & 0.27          & 27.27          & 386.26          & 0.70          & 0.11          & 0.16          & 43.06          & 411.66          \\
                           & DADiff(Ours)            & \textbf{0.63} & \textbf{0.12} & \textbf{0.18} & \textbf{32.15} & \textbf{413.28} & \textbf{0.78} & \textbf{0.11} & \textbf{0.12} & \textbf{33.54} & \textbf{419.26} & \textbf{0.72} & \textbf{0.10} & \textbf{0.14} & \textbf{44.43} & \textbf{424.20} \\ \hline
\end{tabular}
}
\end{table*}

\section{More Visualization Results}

\begin{figure}[htbp]
    \centering
    \includegraphics[width=\linewidth]{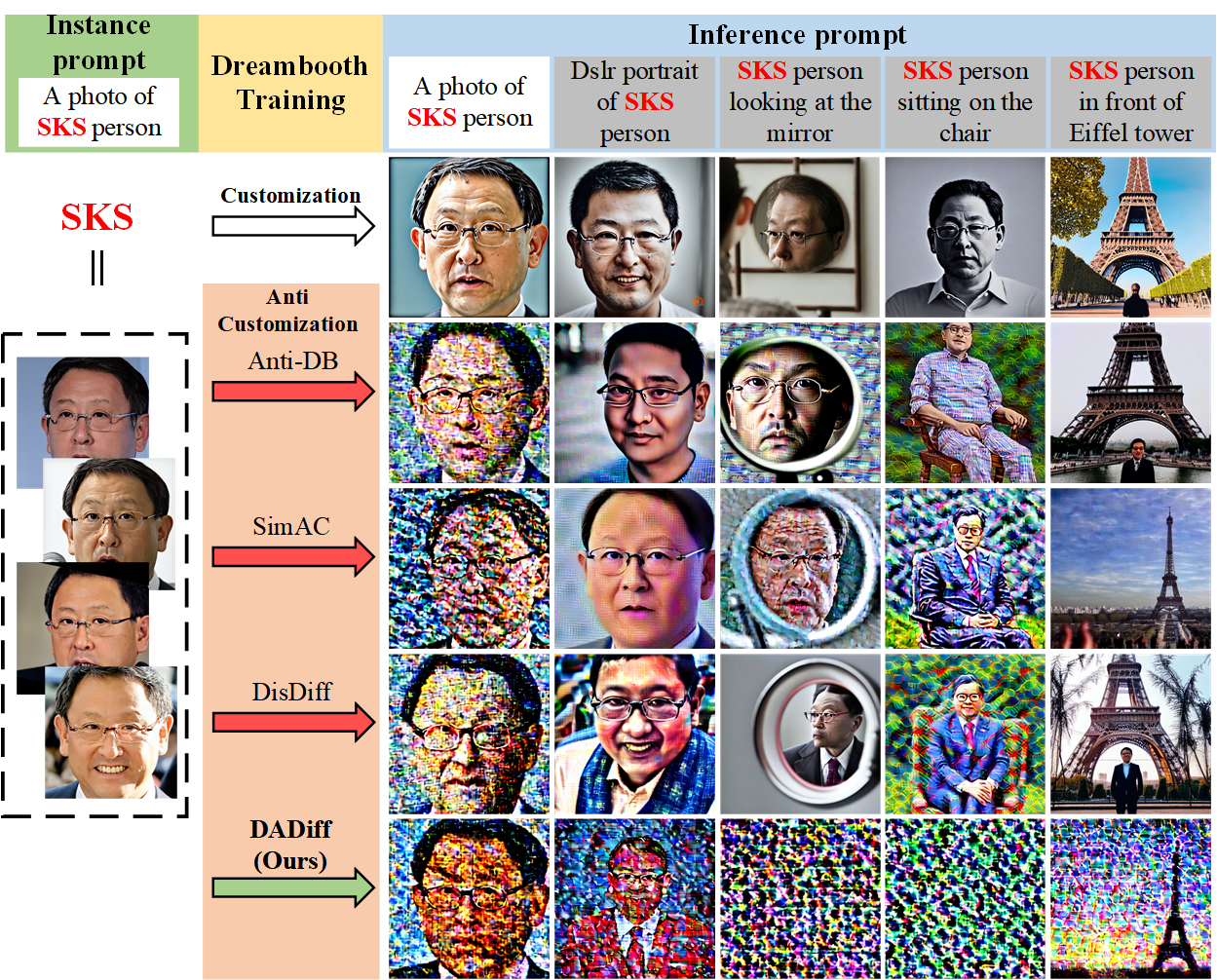}
    \caption{More qualitative evaluations on different IDs from the VGGFace2 dataset.}
\label{fulu_vggface2}
\end{figure}

\begin{figure}[htbp]
    \centering
    \includegraphics[width=\linewidth]{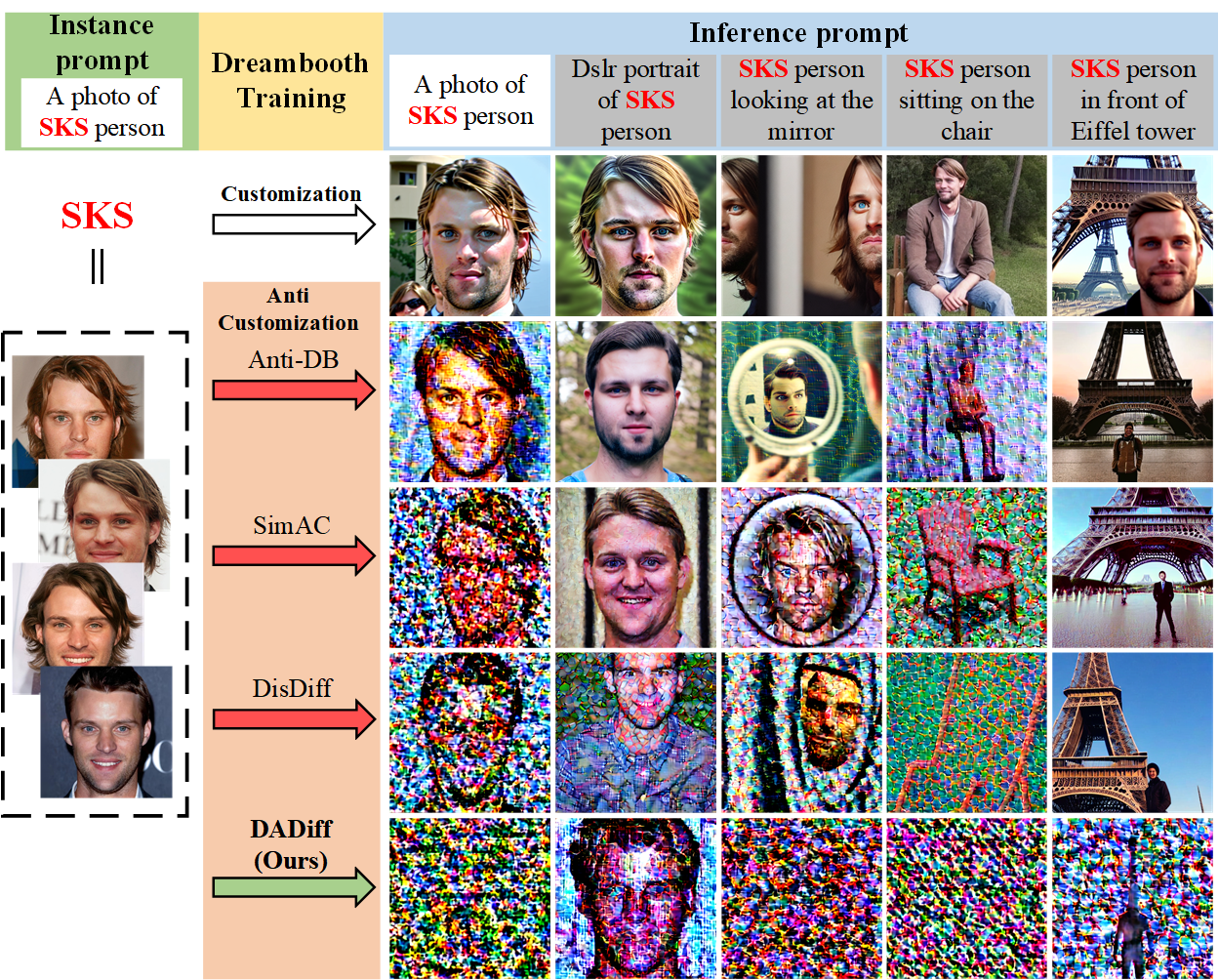}
    \caption{More qualitative evaluations on different IDs from the CelebA-HQ dataset.}
\label{fulu_celebahq}
\end{figure}

In the appendix, we visualize the main text's conducted quantitative experiments to demonstrate the advantages of DADiff more intuitively. We first select more IDs from VGGFaces2 and CelebA-HQ datasets to illustrate the customization of diffusion models and the visualization of generated images for several anti-customization methods. The results are shown in Fig. \ref{fulu_vggface2} and Fig. \ref{fulu_celebahq}. It can be seen that DADiff has achieved excellent model anti-customization effects on different ID images across different datasets and prompts. Existing methods struggle with the black-box inference prompts "\textit{a dslr portrait of sks person}" and "\textit{a photo of sks person in front of Eiffel Tower}", but DADiff still achieves significant results, not only making it difficult to discern the faces in the generated images but also significantly disrupting the texture of the background semantics.

\begin{figure}[htbp]
    \centering
    \includegraphics[width=\linewidth]{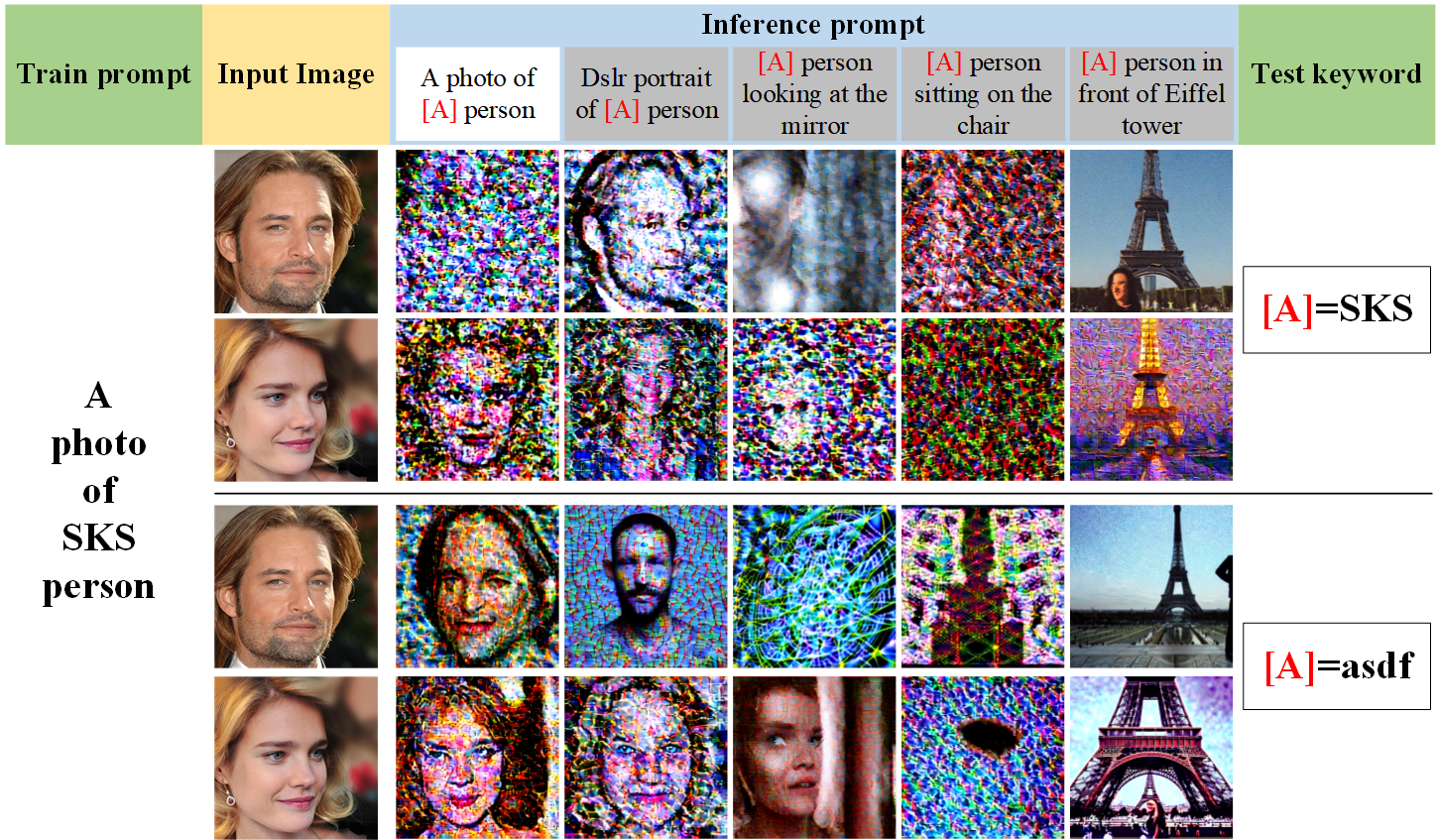}
    \caption{Qualitative evaluations on keyword mismatch cases. The adversarial examples are trained with prompt \textit{"a photo of sks person"}, while the Dreambooth training uses prompt \textit{"a photo of [A] person"}. For the two rows above, \textit{[A]=sks}. For the two rows below, \textit{[A]=asdf}.}
\label{fulu_keyword}
\end{figure}

\begin{figure}[htbp]
    \centering
    \includegraphics[width=\linewidth]{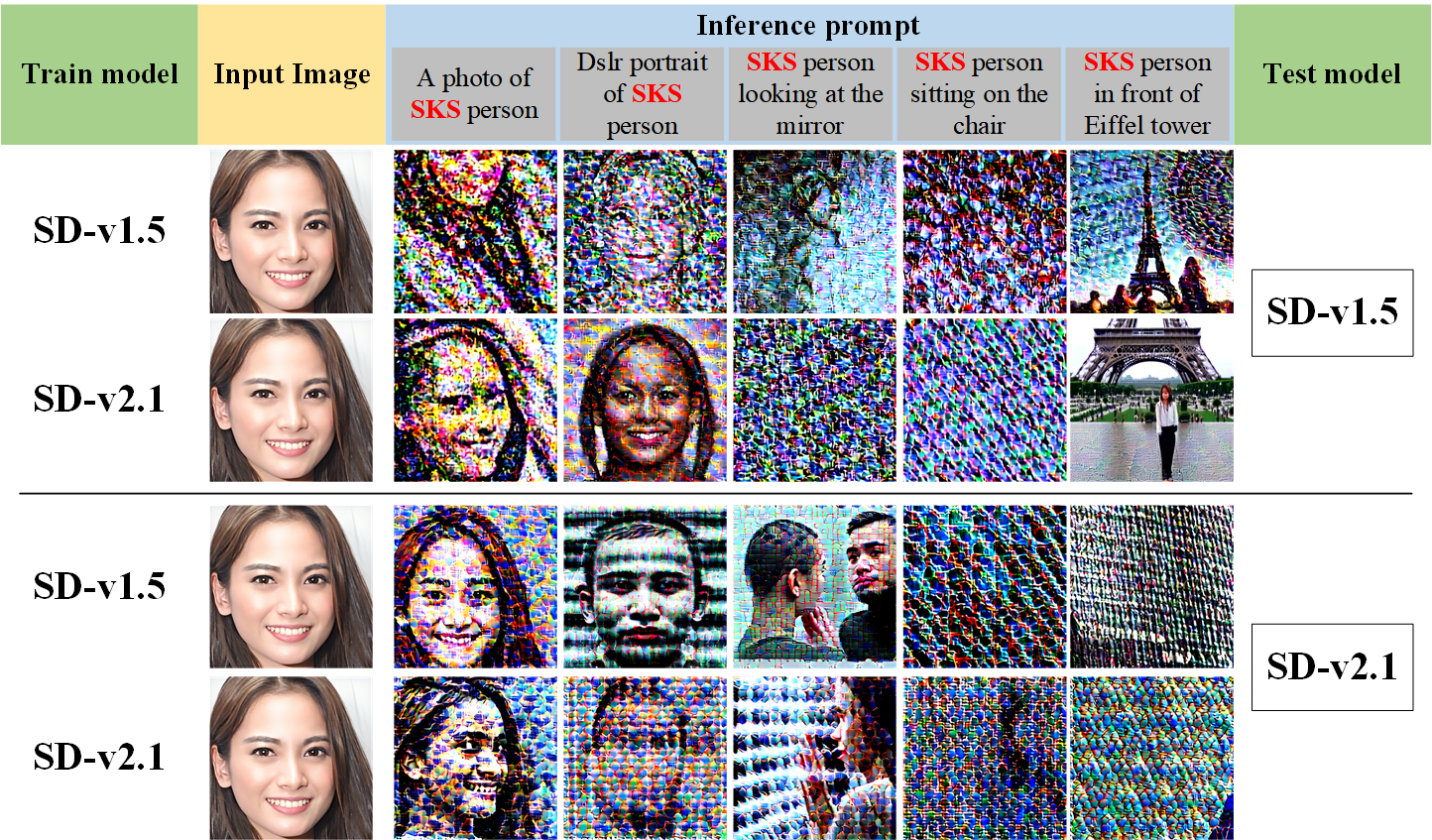}
    \caption{Qualitative visualizations on cross-model evaluations. "Train model" means which model the adversarial examples are trained on, and "Test model" means which model is used for Dreambooth training.}
\label{fulu_model}
\end{figure}

We visualize the anti-customization effects of DADiff in cases of keyword mismatch and cross-model scenarios using images of different facial IDs in Fig. \ref{fulu_keyword} and Fig. \ref{fulu_model}.

In the case of keyword mismatch, although the adversarial effects decline somewhat under certain prompts, DADiff consistently maintains good anti-customization effectiveness. This is particularly evident under the prompts "\textit{a photo of [A] person sitting on the chair}" and "\textit{a photo of [A] person in front of Eiffel Tower}". Even though the adversarial examples are generated using the prompt "\textit{a photo of sks person}," clear images still cannot be generated when "\textit{a photo of asdf person}" is used for Dreambooth training.

In the cross-model scenario, DADiff still exhibits effectiveness, but the destruction effect is weaker compared to the case of keyword mismatch. This indicates that differences in model structure have a greater impact on the performance of adversarial examples than differences at the prompt level. Therefore, attacking different modules of the UNet model effectively improves the cross-model transferability of adversarial examples for diffusion models.

\begin{figure}[htbp]
    \centering
    \includegraphics[width=\linewidth]{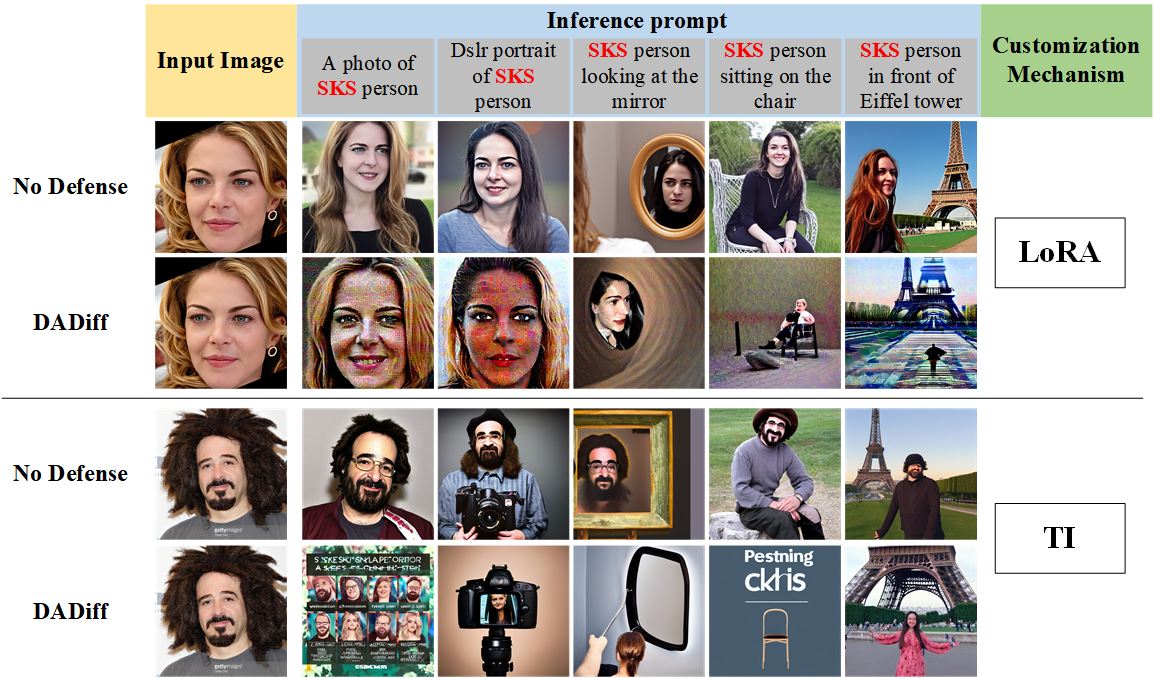}
    \caption{Qualitative visualizations on cross-mechanism evaluations. The adversarial examples are all trained with the Dreambooth mechanism, and "Customization mechanism" means which method is used for customization.}
\label{fulu_mechanism}
\end{figure}

Finally, we visualize the experiment for evaluating the cross-mechanism effect in Fig. \ref{fulu_mechanism}. The results indicate that adversarial examples generated on Dreambooth degrade the visual quality of images customized by LoRA. For cases where the prompt semantics deviate more from the instance prompt ("\textit{a photo of sks person sitting on the chair}" and "\textit{a photo of sks person in front of the Eiffel Tower}"), LoRA fails to generate images containing the correct person ID through adversarial examples. When adversarial examples are used for TI customization training, the customized model is unable to generate images with correct person IDs under various prompts, demonstrating DADiff's ability to counter customization across mechanisms. However, despite achieving the current state-of-the-art, DADiff's disruption effect on cross-mechanism image quality is not sufficiently strong, as also reflected in Table 4 in the main text. This suggests that there are still challenges in the transferability of adversarial examples between different customization strategies, necessitating deeper exploration of the underlying common mechanisms of diffusion model customization techniques in future work.

\section{Ablation Studies}
\subsection{Evaluations between Different Losses}

\begin{table}[hptb]
\centering
\caption{Ablation Studies between using different losses.}
\label{fulutable-ablation1}
\resizebox{\linewidth}{!}{
\begin{tabular}{cccccc}
\hline
\multirow{2}{*}{Method} & \multicolumn{5}{c}{Average}                                                      \\ \cline{2-6} 
                        & FDFR↑         & ISM↓          & SER-FQA↓      & BRISQUE↑       & FID↑            \\ \hline
$L_{cond}$ only              & 0.59          & 0.09          & 0.20          & 34.21          & 430.10          \\
$L_{A}$ only                 & 0.22          & 0.23          & 0.41          & 28.36          & 355.61          \\
$L_{cond}+L_{SA}$               & 0.81          & 0.08          & 0.11          & 34.86          & 451.48          \\
$L_{cond}+L_{CA}$               & 0.66          & 0.09          & 0.16          & 33.90          & 449.65          \\
\textbf{DADiff}         & \textbf{0.93} & \textbf{0.06} & \textbf{0.03} & \textbf{40.59} & \textbf{464.56} \\ \hline
\end{tabular}
}
\end{table}

DADiff consists of several losses, the UNet output loss $L_{cond}$ (Eq. (2) in the main text) and the attention loss $L_{A}$ (Eq. (10) in the main text), where $L_{A}$ includes self-attention loss $L_{SA}$ (Eq. (8) in the main text) and cross-attention loss $L_{CA}$ (Eq. (9) in the main text). We construct ablation experiments for all losses in Table \ref{fulutable-ablation1}, where the values represent the average evaluations of generated images of 32 IDs from two datasets under five inference prompts, similar to Table \ref{fulutable-model}. When only one attention loss is used, its corresponding hyperparameter ($\alpha_{1}$ or $1-\alpha_{1}$) is set to 1. When the gradients of the UNet output loss and any attention loss are combined, the hyperparameter $\alpha_{2}$ is always set to 0.4.

The results demonstrate that the $L_{cond}$ is indeed the primary loss for anti-customization. Excluding $L_{cond}$ leads to a significant decrease in the effectiveness of anti-customization. Adding $L_{SA}$ and $L_{CA}$ to $L_{cond}$ both enhances the perturbation effect of anti-customized adversarial examples. Calculating the loss for the self-attention module makes generated images' faces more unrecognizable (higher FDFR), indicating that disrupting the self-attention module disrupts the custom model's ability to learn pixel correlations. Furthermore, adding the cross-attention module loss on top of this can further degrade the quality of the generated images (lower SER-FQA, higher BRISQUE and FID), suggesting that the introduced text-level adversarial priors can guide further enhancement of the adversarial properties of image-level adversarial examples. In summary, all proposed losses are effective.

\subsection{Single Step vs. LRTGE}

\begin{table}[hptb]
\centering
\caption{Ablation Studies between different gradient update strategies.}
\label{fulutable-ablation2}
\resizebox{\linewidth}{!}{
\begin{tabular}{cccccc}
\hline
\multirow{2}{*}{Method} & \multicolumn{5}{c}{Average}                                                      \\ \cline{2-6} 
                        & FDFR↑         & ISM↓          & SER-FQA↓      & BRISQUE↑       & FID↑            \\ \hline
Single Step             & 0.32          & 0.19          & 0.31          & 25.03          & 372.65          \\
LRTGE                   & \textbf{0.93} & \textbf{0.06} & \textbf{0.03} & \textbf{40.59} & \textbf{464.56} \\ \hline
\end{tabular}
}
\end{table}

DADiff also proposes using the Local Random Timestep Gradient Ensemble (LRTGE) strategy to integrate gradients computed over multiple timesteps and then employs the integrated gradient for perturbation updates in adversarial example generation using PGD attacks. We compare using Single Step update with using LRTGE in the context of considering all losses. The results are presented in Table \ref{fulutable-ablation2}. Due to the absence of considering optimal timestep selection (SimAC) and allocating learning rates based on timesteps (DisDiff) in the ablation experiment, the performance under the Single Step setting is only comparable to that of Anti-DB. This indicates that the gradient values computed at a single timestep are insufficient to encapsulate all the adversarial information contained in the losses, and significantly varying gradient scores and losses indeed hinder the improvement of adversarial example performance. In contrast, using LRTGE to integrate the random step gradient of each time segment can enable a single PGD attack to contain richer adversarial information in the gradient. The fixed time segmentation can also stabilize the overall score of the fused gradient, ultimately significantly improving the effectiveness of the generated adversarial examples.

\end{document}